\documentclass{article}

\usepackage{arxiv}

\usepackage[utf8]{inputenc} 
\usepackage[T1]{fontenc}    
\usepackage{hyperref}       
\usepackage{url}            
\usepackage{booktabs}       
\usepackage{amsfonts}       
\usepackage{nicefrac}       
\usepackage{microtype}      
\usepackage{xcolor}         
\usepackage{amsmath,amssymb,amsthm}
\usepackage{mathtools}
\usepackage{xspace}
\usepackage{epsfig}
\usepackage[lined,noend,linesnumbered,ruled,titlenotnumbered]{algorithm2e}
\usepackage{color}
\usepackage{url}
\usepackage{tikz}
\usepackage{float}
\usepackage{booktabs}
\usepackage{makecell}
\usepackage{multicol}
\usepackage{multirow}
\usepackage{longtable}
\usepackage{array}
\usepackage[libertine]{newtxmath}
\usepackage{colortbl,soul,xcolor,bbm}
\usepackage{subcaption}
\usepackage{nicefrac}
\usepackage{microtype}      
\usepackage[inline]{enumitem}
\usepackage{etoolbox}
\usepackage{listings}

\setlist[itemize]{align=parleft,left=2pt..1.25em}
\SetKwInput{Input}{Inputs}
\SetKwComment{tcc}{//}{}

\SetCommentSty{mycommfont}
\DeclareMathOperator*{\argmax}{arg\!\max}

\DeclareMathOperator*{\avg}{\!avg}
\SetInd{0.3em}{0.5em}
\makeatletter
\patchcmd{\@algocf@start}
  {-1.5em}
  {0pt}
  {}{}
\makeatother

\title{Evolutionary Multi-Objective Diversity Optimization}
\date{}
\author{
Anh Viet Do\and
Mingyu Guo\and
Aneta Neumann\And
Frank Neumann
\affiliations
Optimisation and Logistics, School of Computer and Mathematical Sciences, The University of Adelaide, Adelaide, Australia
}

\begin{document}

\maketitle

\begin{abstract}
Creating diverse sets of high quality solutions has become an important problem in recent years. Previous works on diverse solutions problems consider solutions' objective quality and diversity where one is regarded as the optimization goal and the other as the constraint. In this paper, we treat this problem as a bi-objective optimization problem, which is to obtain a range of quality-diversity trade-offs. To address this problem, we frame the evolutionary process as evolving a population of populations, and present a suitable general implementation scheme that is compatible with existing evolutionary multi-objective search methods. We realize the scheme in NSGA-II and SPEA2, and test the methods on various instances of maximum coverage, maximum cut and minimum vertex cover problems. The resulting non-dominated populations exhibit rich qualitative features, giving insights into the optimization instances and the quality-diversity trade-offs they induce.
\end{abstract}

\section{Introduction}
Diverse solutions problems, seeking multiple maximally distinct solutions of high quality instead of a single solution, have been studied for several decades \cite{Glover2000,Emmanuel05,Ingmar2020,Baste2020,hanaka2020finding,Fomin2021,Fomin20211,hanaka2021,hanaka2022}. They aim to fill the gaps in practical considerations left by traditional optimization. A set of diverse solutions provides robustness in order to deal with changes in the problems, which necessitate changes in current solutions. It also gives the users the choices to address the gaps between the problem models and real-world settings, frequently seen in complex applications with factors that are hard to define or measure \cite{Schittekat2009}. Furthermore, diverse solution sets give the decision makers rich information about the problem instance by virtue of being diverse, which helps augment decision making capabilities. While there are methods to enumerate high quality solutions, having too many overwhelms the decision makers \cite{Glover2000}, and a small, diverse subset can be more useful. It is also known that k-best enumeration tends to yield highly similar solutions, motivating the use of diversification mechanisms \cite{Wang2013,Yuan2015,Hao2020}.

The diverse solutions problems have been studied as an extension to many important and difficult problems. These include constraint satisfaction and optimization problems \cite{Emmanuel05,Petit15,Ruffini2019}, SAT and answer set problem \cite{Nadel2011,Eiter2009}, and mixed integer programming \cite{Glover2000,Danna,Trapp2015}. Recent fixed-parameter tractable algorithms for various graph-based vertex problems \cite{Baste2020} inspired considerations of other combinatorial structures such as trees, paths \cite{hanaka2020finding,hanaka2021}, matching \cite{Fomin20211}, independent sets \cite{Fomin2021}, and linear orders \cite{Arrighi2021}. Furthermore, general frameworks have been proposed for diverse solutions to any combinatorial problem \cite{Ingmar2020,hanaka2022}.

This area of research manifested in Evolutionary Computation literature as \emph{Evolutionary diversity optimization} (EDO). The idea was investigated as early as in the work of Ronald \cite{Ronald} and Zechman and Ranjithan \cite{Zechman2004,Zechman2007}, motivated by practical concerns in real-world problems. The topic was then studied by Ulrich et. al. with more focus on conceptual frameworks \cite{Ulrich2010,Ulrich2011} and has subsequently gained significant attention within the evolutionary computation community. This represents a shift in perspective on diversity, from a necessity in evolutionary search to an optimization goal. Incidentally, around the same time, aspects of evolutionary searches other than diversity were also investigated to address issues that motivated EDO, such as high-performance regions \cite{Parmee2002} and solution's robustness \cite{Tsutsui1997,Branke1998}.

Studies on EDO typically involve defining a search space $S$, an objective function $f$ (to be maximized), a quality threshold $T$, a diversity measure $d$, an integer $r$ and applying evolutionary algorithms to solve the optimization problem
\begin{align}\label{eq:edo_old}
\max_{P\in 2^S:|P|=r}d(P),\quad s.t.\quad\forall x\in P,f(x)\geq T.
\end{align}
Under this paradigm, evolutionary techniques have been investigated in computing diverse Traveling Salesperson Problem (TSP) solutions \cite{Nikfarjam2021,Nikfarjam20211,Do2022}, knapsack packings \cite{Bossek2021}, minimum spanning trees \cite{Bossek20211}, and submodular optimization solutions \cite{Neumann2021}. On the other hand, EDO has seen application in generating images with varying features \cite{Alexander2017}, or to compute diverse TSP instances \cite{Gao2020,Bossek2019} useful for automated algorithm selection and configuration \cite{Kerschke2019}. Different indicators for measuring the diversity of sets of solutions in such as the star discrepancy~\cite{Neumann2018} or those from the area of evolutionary multi-objective optimization~\cite{Neumann2019} have been considered in this paradigm as well.

In this work, we address the treatment of objective quality and diversity of solutions as equal optimization goals. In practice, users may not have enough information about the problem instance in order to formulate quality or diversity criteria that would lead to meaningful optimization outcomes. We re-frame the problem as finding a range of quality-diversity trade-offs which hopefully provides such information to an extent, as well as giving a collection of diverse solution sets to choose from. This approach involves an atypical way of looking at the evolutionary process: each population (i.e. set of solutions) is a unit of evolution, with its own fitness. We describe a basic implementation scheme under this paradigm, which can be realized directly in existing evolutionary algorithms. Concrete examples of its realization are given using NSGA-II and SPEA2 in finding diverse solutions to maximum cut, maximum coverage and minimum vertex cover instances. Experimental investigations give trade-offs of highly varied natures across problems, revealing interesting characteristics of the objective landscapes. Furthermore, these results indicate that the diverse solutions problems can be reasonably addressed under the proposed bi-objective optimization paradigm.

This paper is structured as follows. We include the bi-objective optimization formulation in Section \ref{sec:pre}. The implementation scheme is described in Section \ref{sec:imp}, which is realized in the experimental investigations detailed in Section \ref{sec:exp}. We conclude the paper in Section \ref{sec:con}.

\section{Evolutionary Multi-Objective Diversity Optimization}
\label{sec:pre}
Given an objective function $f$ over a space of feasible solution $S$, the classical optimization problem is specified with $\max_{x\in S}f(x)$.\footnote{This and subsequent formulation also apply to minimization, with trivial differences.} In addition, given a diversity measure $d$, an set aggregating function $F$, and integer $r$, we consider the following bi-objective problem
\begin{align*}
\max_{P\in 2^S:|P|=r}f_1(P):=F\{f(x):x\in P\},f_2(P):=d(P).
\end{align*}
Intuitively, the problem asks to find a set of $r$ solutions representing the best trade-off between diversity and aggregated objective quality. Note this formulation differs from the existing EDO paradigm (Eq. \eqref{eq:edo_old}) in that it does not involve a quality threshold. This makes our formulation suitable for opaque instances.

We consider two aggregating functions in this work: minimum ($F:=\min$) and average ($F:=\avg$). The former consideration models the worst-case robustness requirement on a solution set, and the latter relaxes this requirement to average-case. As for diversity, we look at distance sum measure, frequently considered in dispersion problems \cite{Wang1988,Erkut1990} and recent works on diverse solutions \cite{Baste2020,hanaka2020finding,hanaka2021,hanaka2022}. We give precise definitions in later sections, as it depends on the underlying optimization problem (as specified by $f$ and $S$).

As with many multi-objective problems, the actual computational goal is to find a set of non-dominated trade-offs w.r.t. the objectives. Let $f_i$ be the $i$-th objective function, we say solution $x$ dominates solution $y$ ($x\succeq y$) if $\min_i\{f_i(x)-f_i(y)\}\geq0$, and strong dominance occurs ($x\succ y$) if, in addition, $\max_i\{f_i(x)-f_i(y)\}>0$. Thus, a set of solutions is non-dominated if it contains no pair that exhibits dominance relation.

The problem requires searching in the population space $2^S$ rather than the solution space $S$. Most EAs evolve a population of solutions, whereas we are interested in evolving a population of ``populations''. As such, for the rest of this paper, we modify the classical terminologies in literature to fit the context of the problem. We use ``solution'' to refer to an element $x\in S$, whose quality is $f(x)$. Furthermore, ``individual'' refers to a set of solutions, i.e. an element $I\in 2^S$, and ``population'' refers to a set of individuals\footnote{``Set'' is used loosely: they are technically multisets as self-avoidance is not enforced.}.
\section{Implementing Populations as Individuals}\label{sec:imp}
As the problem we consider is fundamentally a bi-objective optimization problem, applying existing bi-objective optimizers would be, in principle, sufficient in solving it. In this section, we discuss several considerations regarding the implementation of such algorithms that are unique to this problem.
\subsection{Individual representation}\label{sec:rep}
In typical EAs, a population is evolved via the selection-variation-replacement paradigm. However, in this work, we treat populations as individuals, so as to apply standard variation operations on the entire set of solutions directly, effectively performing the 3-step procedure in one 1-step procedure. This is to both simplify the overall algorithm, and ease restrictions on the variation neighborhood in the population space. As such, an individual should be represented in a way that both encodes the information about the solution set and allows efficient implementation of variation operators.

In this work, we adopt the simple concatenation scheme: let $x_i$ be the representation of the $i$-th solution in the individual $I$, $I$ is represented by $x_1x_2\ldots x_r$. For instance, if $x_i$ is a $n$-length bit string, then $I$ is encoded in a $rn$-length bit string, in which each solution is encoded directly as a substring. Having individual representation mirroring solution representation allows each solution to undergo variation when it is applied to the individual, and in the same manner. For example, performing standard bit mutation on the individual essentially does the same to each solution, with the same bit-flip probability. While the scheme does not apply as is to non-linear\footnote{One could extend this scheme to any solution representation with recursive structure (e.g. tree--subtree, string--substring, vector--subspace projection).} solution representations, it suffices for our applications.
\subsection{Recombination}\label{sec:recom}
We see that the concatenation representation is not unique: the solutions ordering itself encodes redundant information in the context of the problem. While positional-bias-free operations (e.g. standard bit mutation) are unaffected by such an artefact, the same cannot be said for recombination as a means to transmit parents' genetic information to offspring. Ensuring transmission of such redundant information may limit the flexibility of the operation, so it should be removed from the process.

Here, we choose to remove such information prior to recombination by shuffling the solutions' positions in one of the parents. This allows each solution in one parent individual to be recombined with any solution in the other parent individual with equal probability. Note that it is sufficient to do this to one parent individual since the solution-to-solution mapping between the parents is what matters, not the ordering itself. Additionally, this scheme introduces an additive $\Theta(r)$ overhead, which is small compared to the $\Theta(rn)$ time cost of the recombination.
\section{Problem Settings and Experiments}\label{sec:exp}
Here, we realize our proposed scheme into concrete algorithms, and perform experimental investigations in max coverage, max cut and min vertex cover problems, all of which are NP-hard. These combinatorial problems are formulated in a way that affords us insights on achievable diversity and its interplay with solution quality; this eases contextualizing and interpreting the results. We select instances from standard benchmark suites and, as we will see, observe qualitatively varied results across problems.

For these graph problems, a solution is a subset of vertices, and the standard representation is indicator vector as bit-string: each bit corresponds to a vertex and assumes value 1 if and only if it is a member of the subset. Our scheme implies that each individual is represented as a $rn$-length bit-string where $n$ is the number of vertices. Also, we use Hamming distance to compute diversity among solutions within an individual, which is the size of two sets' symmetric difference: $|A\Delta B|$. The sum of Hamming distances among a set of bit-strings can be computed and updated efficiently without calculating pairwise distances\footnote{The former takes $\Theta(rn)$ arithmetic operations, and the latter takes $\Theta(k)$ where $k$ is the number of changed bits.}. For consistency, we use set operators with bit-string notations, e.g. $x\in I$ denotes a solution $x$ being in an individual $I$.

The problems impose upper bounds or lower bounds on the sizes of feasible solutions, which we can use to derive upper bounds on diversity. Let $V$ be a ground set, $n:=|V|$, $S:=\{Z\subseteq V:|Z|\leq b\}$ for some integer $b\geq0$, and the diversity function $d(I):=\sum_{x,y\in I}|x\Delta y|$ (each pair is counted once), then we have
\begin{equation}\label{eq:g}
\max_{P\in 2^S:|P|=r}d(P)=g(n,b,r):=nq(r-q)+m(r-2q-1),
\end{equation}
where $h=\min\{b,n/2\}$, and $m\in[0,n),q$ are integers such that $\lceil r/2\rceil\lceil h\rceil+\lfloor r/2\rfloor\lfloor h\rfloor=qn+m$. Intuitively, $d(I)$ is maximized when the number of 1 (or 0) values in $I$ are as close to being equal across bits as possible, and as close to $r/2$ as possible. The reader can confirm that such a configuration under the restriction specified by $S$ yields the right hand side of Eq. \eqref{eq:g} when plugged into $d$. We note that the restriction imposed by the graph problems can be explicit as in cardinality constraint, and implicit in the form of optimal objective value.

We use the well-known NSGA-II \cite{Deb2002} and SPEA2 \cite{Zitzler2001} algorithms with our proposed scheme and the following setting, unless stated otherwise. Specifically, we use implementations in jMetal 5.11 of these algorithms and their standard components \cite{Durillo2011}, and unlisted parameters are assigned the default values.
\begin{itemize}
\item Initialization: Uniform random
\item Output set size: $r\in\{10,20\}$
\item Population/Offspring pool size: $N=20$
\item Parent selection: Binary tournament
\item Crossover method, rate: Uniform crossover, $80\%$
\item Mutation method, strength: Standard bit mutation\footnote{We find higher mutation strengths yield lower final diversity.}, $\chi=0.5/n$
\item Evaluation budget: $5rnN$
\item Number of independent runs per instance: $20$
\end{itemize}
To handle constraint in max coverage and min vertex cover, we penalize the objectives with violation degree. The constraint violation degree of an individual is the sum of such degrees over its solutions: $C(I):=\sum_{x\in I}C(x)$. We detail the settings in the next sections.

\subsection{Maximum Cut}
\begin{figure*}[t!]
\centering
\includegraphics[width=.9\linewidth]{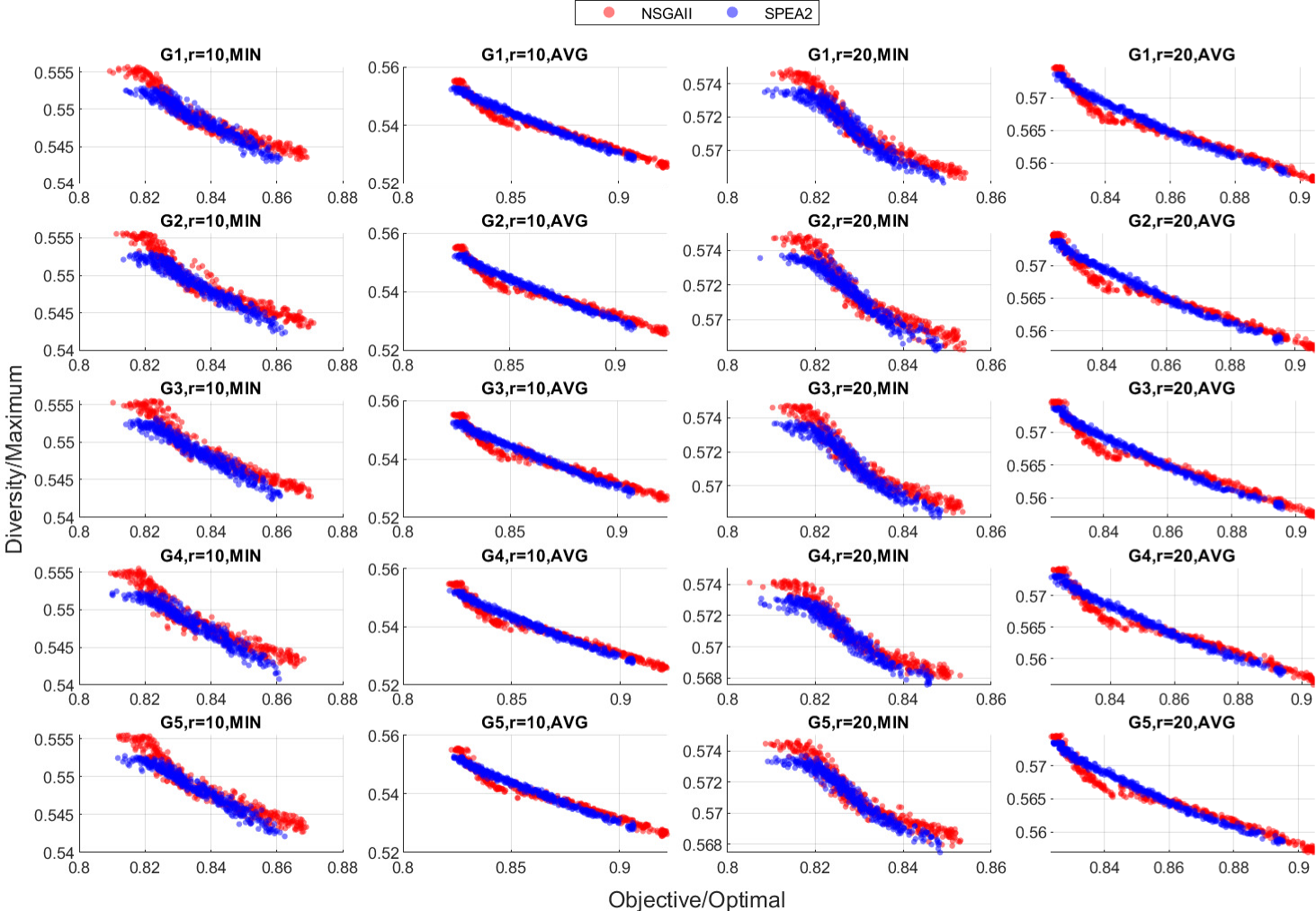}
\caption{Unions of final populations across all runs on Max cut instances. For each run, dominated points are excluded.}\label{fig:tradeoff_maxcut}
\end{figure*}
\begin{table*}[t!]
\centering
\caption{Medians of indicator scores and numbers of non-dominated individuals across runs on Max cut instances.}
\label{tab:results_maxcut}
\scriptsize
\renewcommand{\arraystretch}{1}
\begin{tabular}{cllcccccccccc}\toprule
&\multirow{2}{*}{Inst.}&\multirow{2}{*}{$r$}&\multicolumn{5}{c}{NSGA-II}&\multicolumn{5}{c}{SPEA2}\\\cmidrule(lr){4-13}&&&IGD+&HV&IGD+*&HV*&\#&IGD+&HV&IGD+*&HV*&\#\\\midrule\multirow{10}{*}{\rotatebox[origin=c]{90}{Min objective}}&\multirow{2}{*}{G1}&10&0.49044&\textbf{0.46656}&1.9078e-3&\textbf{0.99578}&20&\textbf{0.49342}&0.45972&\textbf{4.4738e-3}&0.98118&20\\&&20&0.48715&\textbf{0.46116}&1.0426e-3&\textbf{0.9969}&20&\textbf{0.48897}&0.45734&\textbf{2.3626e-3}&0.98864&20\\\cmidrule(lr){2-13}&\multirow{2}{*}{G2}&10&0.48982&\textbf{0.46735}&2.3060e-3&\textbf{0.99428}&20&\textbf{0.49276}&0.46058&\textbf{5.2202e-3}&0.97989&20\\&&20&0.48601&\textbf{0.46213}&1.1482e-3&\textbf{0.99718}&20&\textbf{0.48792}&0.45845&\textbf{2.2096e-3}&0.98923&20\\\cmidrule(lr){2-13}&\multirow{2}{*}{G3}&10&0.49041&\textbf{0.46697}&2.1430e-3&\textbf{0.99505}&20&\textbf{0.49297}&0.46036&\textbf{5.3971e-3}&0.98098&20\\&&20&0.48658&\textbf{0.46151}&1.0371e-3&\textbf{0.99761}&20&\textbf{0.48854}&0.45783&\textbf{2.3923e-3}&0.98966&20\\\cmidrule(lr){2-13}&\multirow{2}{*}{G4}&10&0.49431&\textbf{0.46269}&1.9890e-3&\textbf{0.99442}&20&\textbf{0.49703}&0.4562&\textbf{4.8654e-3}&0.98049&\textbf{20}\\&&20&0.49365&\textbf{0.45447}&1.2396e-3&\textbf{0.99538}&20&\textbf{0.49527}&0.45151&\textbf{2.3533e-3}&0.9889&20\\\cmidrule(lr){2-13}&\multirow{2}{*}{G5}&10&0.49179&\textbf{0.46542}&1.9920e-3&\textbf{0.99573}&20&\textbf{0.49457}&0.45868&\textbf{5.2354e-3}&0.98133&20\\&&20&0.48932&\textbf{0.45888}&1.2962e-3&\textbf{0.99697}&20&\textbf{0.49097}&0.45546&\textbf{2.2058e-3}&0.98954&20\\\midrule\multirow{10}{*}{\rotatebox[origin=c]{90}{Average objective}}&\multirow{2}{*}{G1}&10&0.49325&\textbf{0.49447}&3.2364e-3&\textbf{0.99462}&20&\textbf{0.49434}&0.48454&3.5350e-3&0.97463&\textbf{20}\\&&20&0.48441&\textbf{0.48895}&2.0149e-3&\textbf{0.99706}&20&\textbf{0.48512}&0.48259&\textbf{2.6200e-3}&0.9841&20\\\cmidrule(lr){2-13}&\multirow{2}{*}{G2}&10&0.49237&\textbf{0.49567}&3.5111e-3&\textbf{0.99393}&20&\textbf{0.49372}&0.48534&\textbf{4.3330e-3}&0.97323&20\\&&20&0.4831&\textbf{0.49009}&2.0846e-3&\textbf{0.99641}&20&\textbf{0.48407}&0.48378&\textbf{2.6630e-3}&0.98358&20\\\cmidrule(lr){2-13}&\multirow{2}{*}{G3}&10&0.49265&\textbf{0.49475}&3.0740e-3&\textbf{0.99376}&20&\textbf{0.49398}&0.48461&\textbf{3.9748e-3}&0.9734&20\\&&20&0.48366&\textbf{0.48965}&2.0681e-3&\textbf{0.99707}&20&\textbf{0.48436}&0.48324&\textbf{2.4223e-3}&0.98401&20\\\cmidrule(lr){2-13}&\multirow{2}{*}{G4}&10&0.49668&\textbf{0.49053}&3.0533e-3&\textbf{0.99359}&20&\textbf{0.49812}&0.48078&\textbf{3.9520e-3}&0.97384&20\\&&20&0.49048&\textbf{0.48229}&2.1208e-3&\textbf{0.99644}&20&\textbf{0.49158}&0.47644&\textbf{2.4823e-3}&0.98434&20\\\cmidrule(lr){2-13}&\multirow{2}{*}{G5}&10&0.49413&\textbf{0.49325}&3.0788e-3&\textbf{0.99418}&20&\textbf{0.49542}&0.48336&\textbf{3.9952e-3}&0.97424&20\\&&20&0.48627&\textbf{0.48687}&1.8234e-3&\textbf{0.99703}&20&\textbf{0.48706}&0.48072&\textbf{2.2280e-3}&0.98443&20\\\bottomrule
\end{tabular}
\end{table*}
Given an undirected graph $G=(V,E)$\footnote{Each edge is regarded as a set of two vertices that are its endpoints.}, max cut problem asks to find a solution in
\[\argmax_{x\subseteq V}f(x):=|\{e\in E:|e\cap x|=1\}|.\]
We see that measuring diversity in the vertex space in inappropriate due to $f$ being symmetric. By taking any solution $x$, and duplicating it and $V\setminus x$, we get a population of maximum diversity $g(n,n,r)$ where $g$ is given in Eq. \eqref{eq:g}, and arbitrary quality, effectively reduces the optimal front into a singular point regardless of graph structure. Therefore, we choose to measure diversity in the edge space instead. Let the cut edges of $x$ be $E(x):=\{e\in E:|e\cap x|=1\}$, we have the following fitness functions:
\begin{itemize}
\item Minimum objective setting: $f_1(I):=\min_{x\in I}f(x)$.
\item Average objective setting: $f_1(I):=\frac{1}{r}\sum_{x\in I}f(x)$.
\item $f_2(I):=\sum_{x,y\in I}|E(x)\Delta E(y)|$.
\end{itemize}
The aforementioned insight implies we can upper bound diversity by $g(|E|,OPT,r)$ where $OPT$ is the max cut value. We use the five unweighted instances in G-set benchmark \cite{gset}, containing 800 vertices each, and whose optimal values are known. These values significantly exceed $|E|/2$, meaning for these instances, as cuts in an individual $I$ approach the optimal, more cut edges become over-represented in $I$, diminishing $f_2(I)$. Thus, if the (collective) quality exceeds $|E|/2$, its correlation to diversity is negative, otherwise it is positive. Such conflict between fitnesses should induce rich non-dominated fronts with clear shapes. Note the bound $g(|E|,OPT,r)$ might not be tight since not every edge subset constitutes a cut.

The final populations are visualized in the Fig. \ref{fig:tradeoff_maxcut} and Modified Inverted Generational Distance (IGD+), Hypervolume (HV)\footnote{Smaller IGD+ and greater HV indicate better trade-offs. More comprehensive discussions on performance indicators can be found in \cite{Audet2021}.} and numbers of non-dominated individuals are reported in Table \ref{tab:results_maxcut}. In the table, IGD+ and HV denote values normalized against the extreme objective and diversity values, while IGD+* and HV* are normalized against the best non-dominated fronts aggregated from all runs. Boldface denotes greater medians between two algorithms with statistical significance indicated by Wilcoxon signed-rank tests at 99\% confidence level. The same procedure is used in subsequent experiments.

Overall, the algorithms find many individuals on the non-dominated fronts between 81\% and 92\% of the optimal, consistently across instances. This reveals rich trade-offs, aligning with our intuition regarding quality-diversity correlation. For these instances, 82\% of the optimal cut is approximately $|E|/2$, a point below which we predict the quality-diversity correlation to be positive. This is supported by the shape of the obtained non-dominated fronts within this range, observed more clearly in minimum objective settings.

We see NSGA-II and SPEA2 perform similarly on these instances. Their output non-dominated fronts largely overlap, which small differences in shapes. Inspecting indicators suggests that NSGA-II consistently produce better trade-offs than SPEA2's with statistically significant differences. However, these differences are small in relative magnitude.

We observe that the algorithms consistently reach plateaus before exhausting the budgets. As the Hamming distance sum is easy to maximize, this gives us confidence that the outputs approximate well the high diversity extreme of the optimal fronts. On the other hand, we know there are individuals on the high-quality (low-diversity) end of the fronts that the algorithms fail to approximate as closely. Doing so requires improving the objective quality of all solutions simultaneously. Therefore, expanding the front toward the high-quality end may call for the use of high-performing white-box heuristics.
\subsection{Maximum Coverage}
\begin{figure*}[t!]
\centering
\includegraphics[width=.8\linewidth]{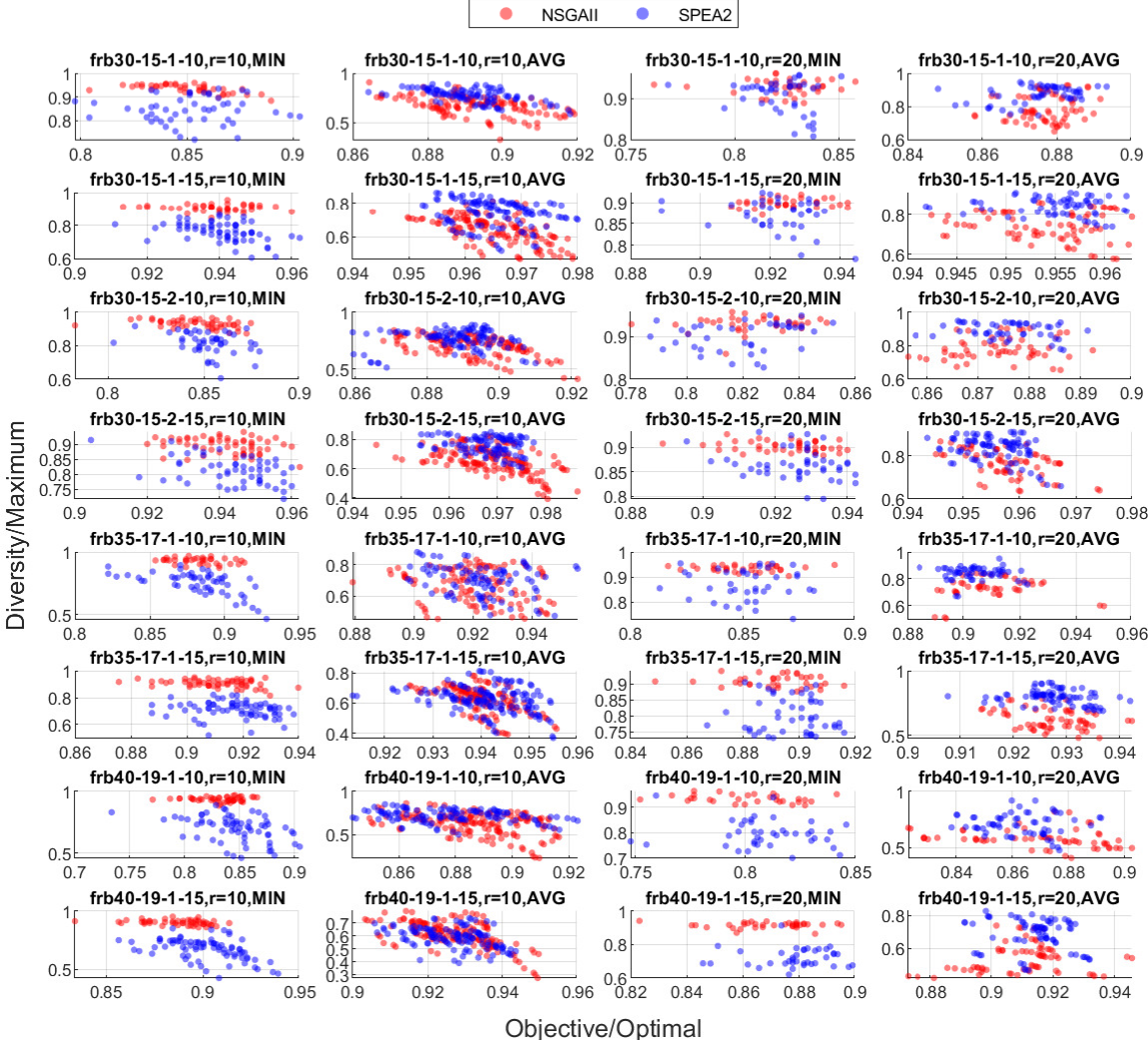}
\caption{Unions of final populations across all runs on Max coverage instances. For each run, dominated points are excluded.}\label{fig:tradeoff_maxcover}
\end{figure*}
\begin{table*}[ht!]
\centering
\caption{Medians of indicator scores and numbers of non-dominated individuals across runs on Max coverage instances.}
\label{tab:results_maxcover}
\scriptsize
\renewcommand{\arraystretch}{1}
\begin{tabular}{cllcccccccccc}\toprule
&\multirow{2}{*}{Inst.}&\multirow{2}{*}{$r$}&\multicolumn{5}{c}{NSGA-II}&\multicolumn{5}{c}{SPEA2}\\\cmidrule(lr){4-13}&&&IGD+&HV&IGD+*&HV*&\#&IGD+&HV&IGD+*&HV*&\#\\\midrule\multirow{16}{*}{\rotatebox[origin=c]{90}{Min objective}}&\multirow{2}{*}{frb30-15-1-10}&10&0.15929&\textbf{0.81026}&0.029592&\textbf{0.93453}&2&\textbf{0.20494}&0.74484&\textbf{0.070245}&0.85907&2\\&&20&0.1832&0.77547&0.031759&0.94084&2&0.19708&0.76149&0.044568&0.92388&2\\\cmidrule(lr){2-13}&\multirow{2}{*}{frb30-15-1-15}&10&0.10914&\textbf{0.86323}&0.025463&\textbf{0.9374}&2&\textbf{0.20583}&0.7614&\textbf{0.11083}&0.82681&3\\&&20&0.12695&\textbf{0.83641}&0.02194&\textbf{0.96042}&1&\textbf{0.14855}&0.80815&\textbf{0.042557}&0.92797&2\\\cmidrule(lr){2-13}&\multirow{2}{*}{frb30-15-2-10}&10&0.16417&\textbf{0.80583}&0.027673&\textbf{0.94845}&2&\textbf{0.21283}&0.73197&\textbf{0.079468}&0.86151&2\\&&20&0.17935&0.7798&0.021634&0.956&2&0.19764&0.76273&\textbf{0.038031}&0.93507&2\\\cmidrule(lr){2-13}&\multirow{2}{*}{frb30-15-2-15}&10&0.10548&\textbf{0.87193}&0.019205&\textbf{0.96324}&2.5&\textbf{0.17837}&0.78967&\textbf{0.079967}&0.87237&2.5\\&&20&0.12098&\textbf{0.84317}&0.019826&\textbf{0.96042}&2&\textbf{0.15161}&0.80867&\textbf{0.048502}&0.92112&2\\\cmidrule(lr){2-13}&\multirow{2}{*}{frb35-17-1-10}&10&0.13118&\textbf{0.8378}&0.03164&\textbf{0.93202}&2&\textbf{0.2302}&0.72143&\textbf{0.10349}&0.80256&2\\&&20&0.15653&\textbf{0.80693}&0.024794&\textbf{0.94912}&1&\textbf{0.19487}&0.74615&\textbf{0.093325}&0.87763&2\\\cmidrule(lr){2-13}&\multirow{2}{*}{frb35-17-1-15}&10&0.12243&\textbf{0.84888}&0.03123&\textbf{0.94851}&3&\textbf{0.2624}&0.69466&\textbf{0.1835}&0.77619&4\\&&20&0.13673&\textbf{0.81768}&0.025891&\textbf{0.94968}&2&\textbf{0.22212}&0.7338&\textbf{0.12012}&0.85226&3\\\cmidrule(lr){2-13}&\multirow{2}{*}{frb40-19-1-10}&10&0.17484&\textbf{0.78967}&0.044668&\textbf{0.90275}&2&\textbf{0.292}&0.67112&\textbf{0.097538}&0.76722&3\\&&20&0.2006&\textbf{0.75525}&0.038091&\textbf{0.92657}&1&\textbf{0.27879}&0.64764&\textbf{0.14262}&0.79455&2\\\cmidrule(lr){2-13}&\multirow{2}{*}{frb40-19-1-15}&10&0.14224&\textbf{0.8172}&0.031981&\textbf{0.91518}&3&\textbf{0.28658}&0.66474&\textbf{0.092133}&0.74444&\textbf{4}\\&&20&0.14977&\textbf{0.80752}&0.021691&\textbf{0.9544}&2&\textbf{0.30992}&0.62822&\textbf{0.19501}&0.74249&2\\\midrule\multirow{16}{*}{\rotatebox[origin=c]{90}{Average objective}}&\multirow{2}{*}{frb30-15-1-10}&10&\textbf{0.27011}&0.67792&\textbf{0.066477}&0.81149&6&0.21687&\textbf{0.73174}&0.035662&\textbf{0.87592}&5\\&&20&\textbf{0.25191}&0.68919&\textbf{0.16121}&0.81487&3&0.16947&\textbf{0.7829}&0.048992&\textbf{0.92568}&2.5\\\cmidrule(lr){2-13}&\multirow{2}{*}{frb30-15-1-15}&10&\textbf{0.28312}&0.69755&\textbf{0.096594}&0.82766&6&0.20168&\textbf{0.77957}&0.027407&\textbf{0.92498}&6\\&&20&\textbf{0.2467}&0.72253&\textbf{0.13882}&0.82469&4&0.15011&\textbf{0.82141}&0.041561&\textbf{0.93754}&3\\\cmidrule(lr){2-13}&\multirow{2}{*}{frb30-15-2-10}&10&0.25897&0.68825&0.049847&0.84212&7&0.21594&0.73583&0.028128&0.90033&6\\&&20&\textbf{0.24879}&0.69083&\textbf{0.13418}&0.8167&3&0.15323&\textbf{0.79726}&0.030719&\textbf{0.94252}&2.5\\\cmidrule(lr){2-13}&\multirow{2}{*}{frb30-15-2-15}&10&\textbf{0.26177}&0.71947&\textbf{0.053042}&0.85601&\textbf{7}&0.20224&\textbf{0.77967}&0.018827&\textbf{0.92764}&5\\&&20&\textbf{0.22441}&0.74949&\textbf{0.050041}&0.84491&3&0.14728&\textbf{0.82496}&0.019408&\textbf{0.93}&\textbf{4}\\\cmidrule(lr){2-13}&\multirow{2}{*}{frb35-17-1-10}&10&0.26252&0.69361&0.095864&0.82278&5&0.2524&0.70518&0.083781&0.83649&5\\&&20&\textbf{0.27384}&0.67771&\textbf{0.091109}&0.75009&3&0.17247&\textbf{0.78145}&0.039561&\textbf{0.86492}&3\\\cmidrule(lr){2-13}&\multirow{2}{*}{frb35-17-1-15}&10&0.31072&0.65554&0.061725&0.84496&6.5&0.32269&0.64837&0.070474&0.83571&6.5\\&&20&\textbf{0.38841}&0.57861&\textbf{0.23436}&0.6746&3&0.20157&\textbf{0.75954}&0.046485&\textbf{0.88555}&4\\\cmidrule(lr){2-13}&\multirow{2}{*}{frb40-19-1-10}&10&0.29465&0.65239&0.069837&0.82183&7&0.26244&0.68673&0.050005&0.86509&5\\&&20&\textbf{0.44666}&0.50248&\textbf{0.23174}&0.60785&3&0.28296&\textbf{0.65226}&0.081098&\textbf{0.78904}&3\\\cmidrule(lr){2-13}&\multirow{2}{*}{frb40-19-1-15}&10&0.3006&0.66013&0.046622&0.87708&7&0.35017&0.61199&0.078987&0.81312&6.5\\&&20&\textbf{0.45212}&0.51614&\textbf{0.20746}&0.65887&3&0.28312&\textbf{0.67128}&0.050441&\textbf{0.85692}&3\\\bottomrule
\end{tabular}
\end{table*}
Given an undirected graph $G=(V,E)$ and threshold $B$, max coverage asks to find a solution in
\[\argmax_{x\subseteq V:|x|\leq B}f(x):=|x\cup\{v\in V:\exists u\in x,\{v,u\}\in E\}|.\]
The constraint violation function on a solution is given by $C(x):=max\{|x|-B,0\}$. We define the fitness functions:
\begin{itemize}
\item Minimum objective setting: $f_1(I):=\min_{X\in I}f(X)$ if $I$ is feasible, $f_1(I):=-C(I)$ otherwise.
\item Average objective setting: $f_1(I):=\frac{1}{r}\sum_{x\in I:C(x)=0}f(x)-C(I)$.
\item $f_2(I):=\sum_{x,y\in I}|x\Delta y|-r|V|C(I)$.
\end{itemize}
We find that the algorithm struggles to maintain feasibility of many solutions simultaneously\footnote{Preliminary runs with standard mutation yield no feasible outputs when $r=20$.}, so we use a modified mutation operator. It flips each 1-bit with probability $\chi(C(x)+1)$, 0-bit with probability $\chi$, where $x$ is the solution it belongs to. This reduces generation of infeasible offspring and mitigates stagnation, with small overhead.

We use the complement of BHOSLIB instances \cite{BHOSLIB}, denoted with \lstinline|{graphname}-{threshold}|. While these graphs are sparse, the coverage functions they induce exhibit high degrees of multimodality, i.e. there are many distant near-optimal local optima. This means we can expect the optimal non-dominated front to be close to $(OPT,g(n,B,r))$, the former being the maximum coverage value and the latter being the diversity upper bound. Note that we can construct a feasible individual $I$ with $f_2(I)=g(n,B,r)$, implying the bound is tight and always meets the optimal front at an individual.

The results are shown in Fig. \ref{fig:tradeoff_maxcover} and Table \ref{tab:results_maxcover}. The algorithms return individuals with high objective values, and relatively small variations along this dimension. This leads to fewer non-dominated individuals from each run, as only few fitness values are occupied. Furthermore, most output individuals have high diversity, reaching above 80\% of the upper bound in many instances. High Hypervolumes and small Inverted Generational Distances against $(OPT,g(n,B,r))$ indicate that the output non-dominated fronts occupy a large part of the fitness space; this implies that the optimal front is close to $(OPT,g(n,B,r))$, as we suspect.

The two algorithms seem to perform similarly on these instances. The fronts produced by NSGA-II have slightly higher diversity than those by SPEA2 in the minimum objective settings, while the rest see large overlaps. The achieved indicators show mixed comparisons. In min objective setting, NSGA-II achieve better non-dominated fronts in most cases. In average objective setting, SPEA2 reaches better indicator scores, at greater frequency when $r=20$.

\subsection{Minimum Vertex Cover}
Given an undirected graph $G=(V,E)$, min vertex cover asks to find a solution in
\[\argmax_{x\subseteq V}f(x):=|V\setminus x|\quad s.t.\quad\forall e\in E,e\cap x\neq\emptyset.\]
The constraint violation function on a solution is the number of edges not covered $C(x):=|\{e\in E:e\cap x=\emptyset\}|$. We define the fitness functions:
\begin{itemize}
\item Minimum objective setting: $f_1(I):=\min_{X\in I}f(x)$ if $I$ is feasible, $f_1(I):=-C(I)$ otherwise.
\item Average objective setting: $f_1(I):=\frac{1}{r}\sum_{x\in I:C(x)=0}f(x)-C(I)$.
\item $f_2(I):=\sum_{x,y\in I}|x\Delta y|-r|E|C(I)$.
\end{itemize}
In addition, we implement a simple repair heuristic used in \cite{Pelikan2007}. After mutation, infeasible solutions are repaired with the procedure outlined in Algorithm \ref{alg:repair_mvc}. This incurs an additive $\Theta(|E|)$ overhead per infeasible solution, similar to the time cost of feasibility checking\footnote{The
cost is further reduced by streamlining repairing with evaluation.}. When this heuristic is used, $C(I)$ is not computed.
\begin{algorithm}[t!]
\KwIn{$G=(V,E)$, $x\subseteq V$}
\KwOut{$x$}
\For{0-bit $i$ in $x$ in random order}{
\If{$i$ is incident to an edge not covered by $x$}{
$x_i\gets1$, update covered edges\;
}
}
\For{1-bit $j$ in $x$ in random order}{
\If{$j$ is adjacent to no vertex in $V\setminus x$}{
$x_j\gets0$\;
}
}
\caption{Repair heuristic for Min vertex cover \cite{Pelikan2007}}
\label{alg:repair_mvc}
\end{algorithm}
We choose nine hard BHOSLIB instances \cite{BHOSLIB} and complement of three DIMACS instances for maximum clique problem \cite{dimacs}. BHOSLIB instances contain only large vertex covers, so maximizing objective correlates with maximizing diversity as both involves minimizing common selected vertices. This also means there are few possible objective values among feasible solutions, so we can expect the non-dominated fronts returned by the algorithms to be sparse. Meanwhile, selected DIMACS graphs are regular, inducing diverse sets of minimum vertex covers. This means we can expect the optimal non-dominated fronts in these instances to be close to $(OPT,g(n,OPT,r))$ where $OPT$ is the optimal objective value.

\begin{figure*}[t!]
\centering
\includegraphics[width=1\linewidth]{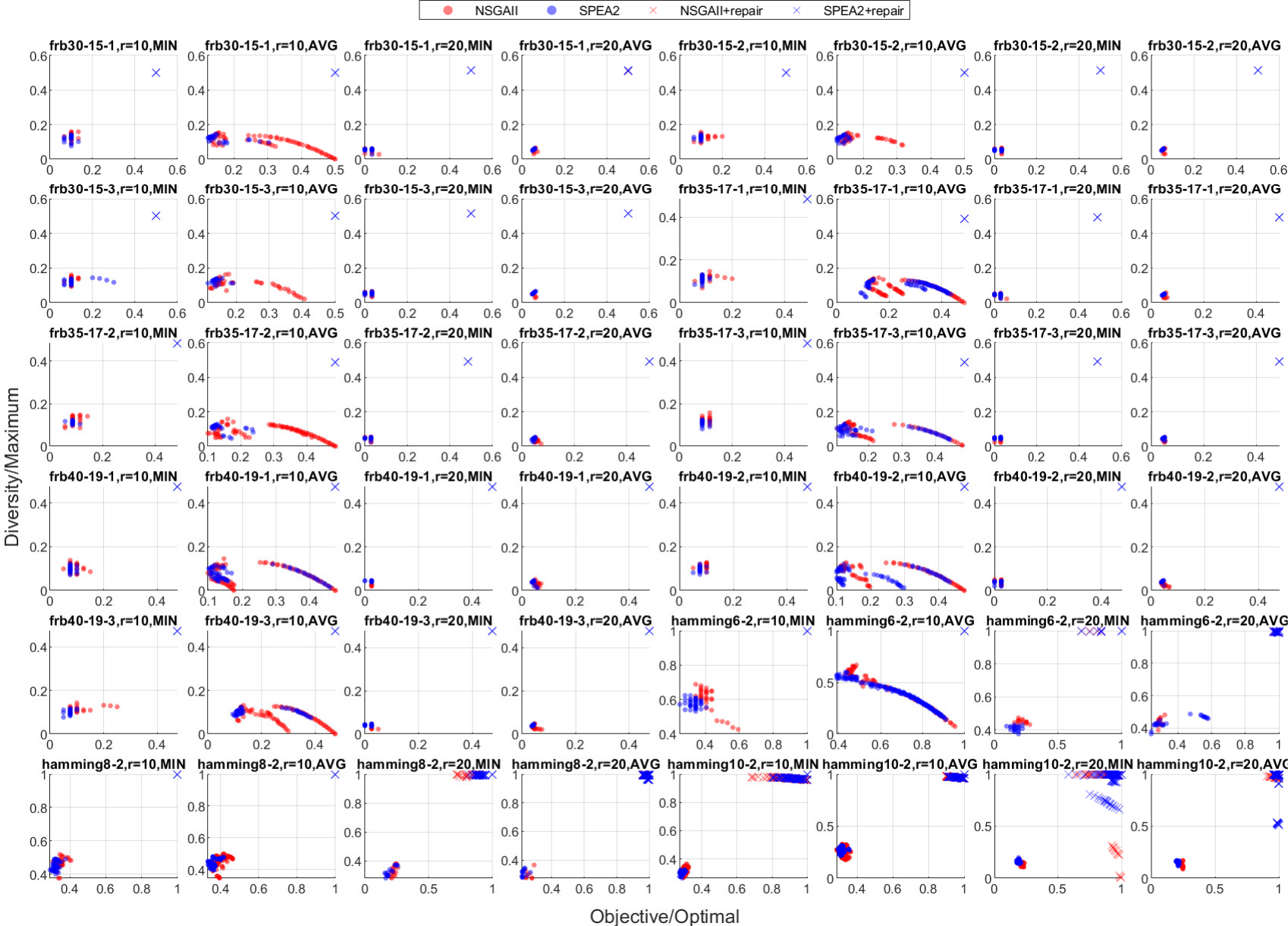}
\caption{Unions of final populations across all runs on Min vertex cover instances. For each run, dominated points are excluded.}\label{fig:tradeoff_mvc}
\end{figure*}

The results are shown in Fig. \ref{fig:tradeoff_mvc} and Table \ref{tab:results_mvc}. We see both algorithms with the repair heuristic consistently converge at a single non-dominated individual. In BHOSLIB instances, this individual is mapped to approximate 50\% of optimal objective and diversity upper bound, resulting in Hypervolumes at roughly 25\% of the entire fitness space. In DIMACS instances, this individual is mapped to the best theoretical trade-off consisting of $(OPT,g(n,OPT,r))$, agreeing with our intuition. The few number of returned non-dominated individuals could be explained by the limited number of distinct fitness values a feasible individual can assume, combined with the positive correlation between maximizing the two fitnesses. The latter explains the observation that whenever multiple non-dominated individuals are returned, they are all dominated by the one individual the searches seem to converge at in the same instance.

We see NSGA-II and SPEA2 perform similarly under the same configuration across instances. Their achieved indicators are identical in most instances, and no statistically significant difference is detected in the rest. Furthermore, the repair heuristic produces clear improvements over non-repair variants, with roughly 35\% run-time overhead. The non-repair variants also fail to produce comparable trade-offs within 150\% the evaluation budgets.

We observe similar output individuals across instances within the same class (e.g. prefixed by ``frb30-15''), resulting in identical median indicator scores. This indicates the similarity in the objective landscapes induced by these graphs. In addition, the outputs are similar between min objective setting and average objective setting. The choice of $f_1$ does not seem to influence the behaviors of the algorithms on these instances.

\begin{table}[ht!]
\centering
\caption{Medians of indicator scores across runs on Min vertex cover instances. Medians of IGD+*, HV* and numbers of non-dominated points are 1, 0, 0, respectively, for both algorithm in all instances, with no statistically significant differences.}
\label{tab:results_mvc}
\small
\renewcommand{\arraystretch}{.9}
\begin{tabular}{cllcccc}\toprule
&\multirow{2}{*}{Inst.}&\multirow{2}{*}{$r$}&\multicolumn{2}{c}{NSGA-II}&\multicolumn{2}{c}{SPEA2}\\\cmidrule(lr){4-7}&&&IGD+&HV&IGD+&HV\\\midrule\multirow{24}{*}{\rotatebox[origin=c]{90}{Min objective}}&\multirow{2}{*}{frb30-15-1}&10&0.70711&0.25&0.70711&0.25\\&&20&0.69762&0.25676&0.69762&0.25676\\\cmidrule(lr){2-7}&\multirow{2}{*}{frb30-15-2}&10&0.70711&0.25&0.70711&0.25\\&&20&0.69762&0.25676&0.69762&0.25676\\\cmidrule(lr){2-7}&\multirow{2}{*}{frb30-15-3}&10&0.70711&0.25&0.70711&0.25\\&&20&0.69762&0.25676&0.69762&0.25676\\\cmidrule(lr){2-7}&\multirow{2}{*}{frb35-17-1}&10&0.72731&0.23592&0.72731&0.23592\\&&20&0.72182&0.2397&0.72182&0.2397\\\cmidrule(lr){2-7}&\multirow{2}{*}{frb35-17-2}&10&0.72731&0.23592&0.72731&0.23592\\&&20&0.72182&0.2397&0.72182&0.2397\\\cmidrule(lr){2-7}&\multirow{2}{*}{frb35-17-3}&10&0.72731&0.23592&0.72731&0.23592\\&&20&0.72182&0.2397&0.72182&0.2397\\\cmidrule(lr){2-7}&\multirow{2}{*}{frb40-19-1}&10&0.74246&0.22562&0.74246&0.22562\\&&20&0.74069&0.22682&0.74069&0.22682\\\cmidrule(lr){2-7}&\multirow{2}{*}{frb40-19-2}&10&0.74246&0.22562&0.74246&0.22562\\&&20&0.74069&0.22682&0.74069&0.22682\\\cmidrule(lr){2-7}&\multirow{2}{*}{frb40-19-3}&10&0.74246&0.22562&0.74246&0.22562\\&&20&0.74069&0.22682&0.74069&0.22682\\\cmidrule(lr){2-7}&\multirow{2}{*}{hamming6-2}&10&0&1&0&1\\&&20&0&1&0&1\\\cmidrule(lr){2-7}&\multirow{2}{*}{hamming8-2}&10&0&1&0&1\\&&20&0&1&0&1\\\cmidrule(lr){2-7}&\multirow{2}{*}{hamming10-2}&10&0&1&0&1\\&&20&0&1&0&1\\\midrule\multirow{24}{*}{\rotatebox[origin=c]{90}{Average objective}}&\multirow{2}{*}{frb30-15-1}&10&0.70711&0.25&0.70711&0.25\\&&20&0.69762&0.25676&0.69762&0.25676\\\cmidrule(lr){2-7}&\multirow{2}{*}{frb30-15-2}&10&0.70711&0.25&0.70711&0.25\\&&20&0.69762&0.25676&0.69762&0.25676\\\cmidrule(lr){2-7}&\multirow{2}{*}{frb30-15-3}&10&0.70711&0.25&0.70711&0.25\\&&20&0.69762&0.25676&0.69762&0.25676\\\cmidrule(lr){2-7}&\multirow{2}{*}{frb35-17-1}&10&0.72731&0.23592&0.72731&0.23592\\&&20&0.72182&0.2397&0.72182&0.2397\\\cmidrule(lr){2-7}&\multirow{2}{*}{frb35-17-2}&10&0.72731&0.23592&0.72731&0.23592\\&&20&0.72182&0.2397&0.72182&0.2397\\\cmidrule(lr){2-7}&\multirow{2}{*}{frb35-17-3}&10&0.72731&0.23592&0.72731&0.23592\\&&20&0.72182&0.2397&0.72182&0.2397\\\cmidrule(lr){2-7}&\multirow{2}{*}{frb40-19-1}&10&0.74246&0.22562&0.74246&0.22562\\&&20&0.74069&0.22682&0.74069&0.22682\\\cmidrule(lr){2-7}&\multirow{2}{*}{frb40-19-2}&10&0.74246&0.22562&0.74246&0.22562\\&&20&0.74069&0.22682&0.74069&0.22682\\\cmidrule(lr){2-7}&\multirow{2}{*}{frb40-19-3}&10&0.74246&0.22562&0.74246&0.22562\\&&20&0.74069&0.22682&0.74069&0.22682\\\cmidrule(lr){2-7}&\multirow{2}{*}{hamming6-2}&10&0&1&0&1\\&&20&0&1&0&1\\\cmidrule(lr){2-7}&\multirow{2}{*}{hamming8-2}&10&0&1&0&1\\&&20&0&1&0&1\\\cmidrule(lr){2-7}&\multirow{2}{*}{hamming10-2}&10&0&1&0&1\\&&20&0&1&0&1\\\bottomrule
\end{tabular}
\end{table}

\section{Conclusions}\label{sec:con}
In this work, we study a bi-objective optimization formulation of the diverse solutions problem, where different trade-offs between solutions objective quality and diversity are evolved. This formulation requires that the output be a collection of solution sets, in exchange for eliminating the need to set quality or diversity criteria. We present an implementation scheme that treats a set of solution as an individual, and handles the inherent symmetry in diversity measures. We realize the scheme in NSGA-II and SPEA2, and test the methods on various maximum cut, maximum coverage and minimum vertex cover instances. The results reveal insights on the optimization instances to which diverse solutions are computed, and confirm that the bi-objective optimization paradigm can be used to address the diverse solutions problem.

We remark that moving from solution spaces to populations spaces introduces extra complexity to the search processes in a way that seems to frustrate black-box approaches. It appears that this additional difficulty can be overcome by problem-specific heuristics, as observed in our investigation with min vertex cover. We speculate that state-of-the-art for this problem will involve memetic algorithms and hybrid approaches.

\section*{Acknowledgments}
This work has been supported by the Australian Research Council (ARC) through grants DP190103894 and FT200100536.

\bibliographystyle{abbrv}
\bibliography{refs}

\end{document}